\documentclass[english,twoside]{IEEEtran}
\usepackage[T1]{fontenc}
\usepackage[latin9]{inputenc}
\usepackage{amsmath}
\usepackage{amsthm}
\usepackage{amssymb}
\usepackage{graphicx}
\usepackage{xcolor}
\usepackage[ruled,vlined]{algorithm2e}
\usepackage[authoryear]{natbib}
\usepackage{hyperref}

\makeatletter

\theoremstyle{plain}
\newtheorem{thm}{\protect\theoremname}
\theoremstyle{definition}
\newtheorem{example}[thm]{\protect\examplename}

\makeatother

\usepackage{babel}
\providecommand{\examplename}{Example}
\providecommand{\theoremname}{Theorem}

\begin{document}

\title{Meta-learning of Sequential Strategies}

\markboth{DeepMind Tech Report, \today, V1.1}{Meta-learning of Sequential Strategies}

\author{Pedro A.\ Ortega, Jane X.\ Wang, Mark Rowland, Tim Genewein, Zeb Kurth-Nelson, Razvan Pascanu,\\ Nicolas Heess, Joel Veness, Alex Pritzel, Pablo Sprechmann, Siddhant M.\ Jayakumar, Tom McGrath, Kevin Miller, Mohammad Azar, Ian Osband, Neil Rabinowitz, Andr\'as Gy\"orgy, Silvia Chiappa, Simon Osindero,\\ Yee Whye Teh, Hado van Hasselt, 
Nando de~Freitas, Matthew Botvinick, and Shane Legg\\
DeepMind}

\maketitle

\begin{abstract}
In this report we review memory-based meta-learning as a tool for
building sample-efficient strategies that learn from past experience
to adapt to any task within a target class. Our goal is to equip the
reader with the conceptual foundations of this tool for building
new, scalable agents that operate on broad domains.
To do so, we present basic algorithmic templates for building 
near-optimal predictors and reinforcement learners which behave 
as if they had a probabilistic model that allowed them to efficiently 
exploit task structure.
Furthermore, we recast memory-based meta-learning within a Bayesian
framework, showing that the meta-learned strategies are near-optimal
because they amortize Bayes-filtered data, where the adaptation is
implemented in the memory dynamics as a state-machine of 
sufficient statistics. Essentially, memory-based meta-learning 
translates the hard problem of probabilistic sequential
inference into a regression problem.
\end{abstract}


\section{Introduction}

How do we build agents that perform well over a wide range of tasks?
Achieving this generality (or universality) is considered by many
research agendas to be one of the core challenges in artificial intelligence (AI) 
\citep{solomonoff1964formal, gottfredson1997mainstream, 
hernandez2000beyond, hutter2004universal}. Generality
is at the heart of IQ tests \citep{spearman1904general, 
raven1936mental, urbina2011tests} and the measure of 
intelligence suggested by \citet{legg2007universal}. 

\emph{Meta-learning} is a practical machine learning 
approach to building \emph{general AI} systems such 
as classifiers, predictors, and agents. Broadly speaking, 
meta-learning aims to produce flexible, data-efficient learning 
systems through the acquisition of inductive biases from data 
\citep{bengio1991learning, schmidhuber1996simple, thrun1998learning}. 
In contrast to systems that build in such biases by design, in meta-learning 
they are acquired by training on a \emph{distribution over tasks}. 
For example, an agent trained to find rewards in one maze will 
simply learn the solution to that maze, but an agent trained on 
mazes drawn from a broad class will learn a general-purpose 
strategy for exploring new mazes \citep{wang2016learning, duan2016rl}. 
Systems trained in this way are able to absorb structure in the 
task distribution that allows them to adapt efficiently and 
generalize to new tasks, effectively leveraging past experience 
to speed up new learning. Therefore, a  meta-learner can be 
thought of as a primary learning system that progressively 
improves the learning of a secondary system \citep{bengio1991learning,
schmidhuber1996simple, thrun1998learning, hochreiter2001learning}.

This method has been shown to be remarkably effective in practice, 
especially in combination with deep learning architectures. Recent 
years have brought a wealth of approaches centered on meta-learning 
different aspects of the learning process, such as learning the 
optimizer \citep{andrychowicz2016learning, li2016learning, 
ravi2016optimization, wichrowska2017learned, chen2018learning}, 
the metric space \citep{vinyals2016matching, snell2017prototypical}, 
the initial network parameters \citep{finn2017model, nichol2018reptile}, 
the learning targets \citep{Xu2018meta}, conditional distributions 
\citep{wang2017robust, garnelo2018conditional, gordon2018meta,
zintgraf2018caml, chen2018sample}, or even the entire learning procedure 
using a memory-based architecture such as a recurrent neural network 
\citep{santoro2016meta, wang2016learning, duan2016rl, denil2016learning,
mishra2018simple}. Some approaches have also taken advantage of modularity
as an inductive bias to learn modules to be re-used in transfer tasks
\citep{reed2015neural}.

In this report we focus on this last class of \emph{memory-based 
meta-learning methods}, which aim to find sequential strategies 
that learn from experience. 
Specifically, we aim for a theoretical understanding of such 
meta-learning methods by recasting them within a Bayesian framework. 
Our goal is to provide a basic algorithmic template to which various 
meta-learning procedures conform, showing that learned strategies 
are capable of performing near-optimally. We hope that this deeper 
conceptual understanding will provide the foundations for new, 
scalable models.

The significance of memory-based meta-learning methods rests in their 
ability to build---in a scalable and data-driven way---systems that 
behave \emph{as if} they had a probabilistic model of the future. 
Agents with probabilistic models possess inductive biases that allow 
them to quickly draw structured inferences based on experience. However, 
building these agents is very challenging: typically, constructing them 
involves specifying both the probabilistic models and their inference
procedures either by hand or through probabilistic programming. 

In contrast, meta-learning offers a simple alternative: to precondition 
a system with training samples in order to fix the right inductive 
biases necessary at test time.
\emph{The key insight is that the meta-training process 
generates training samples that are implicitly filtered according 
to Bayes rule, i.e.\ the samples are
drawn directly from the posterior predictive distribution}.
Combined with a suitably chosen cost function, meta-learning can
use these Bayes-filtered samples to regress an adaptive strategy that
solves a task quickly by implicitly performing Bayesian updates 
``under the hood''---that is, without computing the 
(typically intractable) Bayesian updates explicitly. In this way,
memory-based meta-learning agents can behave as if they possess a 
probabilistic model \citep{orhan2017efficient}. Moreover, the agents track, in their memory dynamics, 
the Bayesian sufficient statistics necessary for estimating the uncertainties
for solving the task. Note that this conceptualization is distinct 
from but extends ideas presented in \citet{baxter1998theoretical, baxter2000model}, 
which considered only the supervised learning case, and 
\citet{finn2017model, grant2018recasting}, in which the inference 
step is built-in and constrained to only a few gradient-descent steps. 

This report is structured as follows. Throughout, we focus on simple 
toy examples before moving on to discuss issues with scaling and 
practical applications. Section~\ref{sec:seq-pred} reviews
sequential predictions. This is the most basic 
application of meta-learning of sequential strategies,
as it only requires regressing the statistics of the training samples.
The analysis of the sequential prediction case will also serve as
a basis for studying other applications and for investigating
the structure of the solutions found 
through meta-learning. Section~\ref{sec:seq-dm} reviews the sequential
decision-making case. Here we show how to combine the basic meta-learning
scheme with a policy improvement method. We illustrate this with two minimal examples: 
one for building Thompson sampling agents, which is the natural extension 
of the prediction case, and another for building Bayes-optimal agents.
Finally, Section~\ref{sec:discussion} discusses 
the connection between meta-learning
and Bayesian statistics, the spontaneous emergence of meta-learned solutions
in (single-task) online learning, and future challenges.

\section{Sequential Prediction}\label{sec:seq-pred}

We start our analysis with the problem of sequential prediction, i.e.\ 
the task of forecasting the future based on past experience.
We use this case because sequential prediction is the most basic 
application of meta-learning, and it will lay down the basics 
for analyzing other applications such as sequential decision-making.

Consider the following sequence prediction problems:
\begin{description}
\item [{I)}] 1, 0, 1, 1, 0, 1, 1, 1, 0, 1, 1, ?
\item [{II)}] 1, 4, 9, 16, 25, ?
\item [{III)}] 1, 2, 3, 4, ?
\end{description}
What is the next number in the sequence? The answers are given by :
\begin{description}
\item [{I)}] 1, 0, 1, 1, 0, 1, 1, 1, 0, 1, 1, $\mathbf{1}$
\item [{II)}] 1, 4, 9, 16, 25, \emph{$\mathbf{36}$}
\item [{III)}] 1, 2, 3, 4, \emph{$\mathbf{29}$}
\end{description}
These are justified as follows. Sequence~(I) is a unary encoding
of the natural numbers. (II) is the sequence of quadratic numbers
$s(t)=t^{2}$. Finally, (III) is the sequence $s(t)=t^{4}-10t^{3}+35t^{2}-49t+24$.
Intuitively however, 29 feels very unlikely, as~5 (i.e.~the next natural
number) seems a far more natural choice compared to the complex 4th-degree
polynomial \citep{hutter2004universal}. 

In spite of the enormous number of possible explanations, 
we were remarkably good at predicting the next element in the sequence.
Our prediction ability is at the same time general
and data-efficient. We possess inductive 
biases that allowed us to quickly narrow down the
space of possible sequences to just a few, even though there isn't
any obvious shared structure among the three examples; 
furthermore, these biases permitted us to judge the 
relative plausibilities of competing predictions (such as 
``29'' versus ``5'' in the previous example). 
Our prediction strategy appears to follow two principles: 
we maintained every possible explanation/hypothesis we could think of
given resource constraints (\emph{Epicurus' principle}), 
and we deemed simpler explanations
as being more likely (\emph{Occam's razor}). These are also
the core principles underlying Bayesian statistics, which is
arguably the gold standard for computing predictions given the 
inductive biases \citep{jaynes2003probability}.

But where do these inductive biases come from? How does
a learner know what is most parsimonious? Here, meta-learning provides
a simple answer: a learner, through repeated exposure to various
tasks, effectively captures the statistics of the data,
which translate into the inductive biases necessary for
future predictions.

In the following we will briefly review how to make sequential predictions
using Bayesian ideas. Then, we will show how to numerically approximate
this prediction strategy using meta-learning. 

\subsection{Problem setup}

We now formalize the setup for sequential prediction. 
For convenience, we limit ourselves to finite observation 
spaces and discrete time.%
\footnote{Note however that this assumption
comes with a loss of generality. Extensions to e.g.\ continuous
domains typically require additional (geometric) assumptions which
are beyond the scope of this report.}
Outputs are distributions over finite spaces unless stated otherwise.

Our goal is to set up a generative process over trajectories (i.e.\ finite
sequences of observations), where the trajectories are drawn not from
a single generator, but from a class of generators. Each generator will
correspond to a possible ``ground truth'' that we want the system to
consider as a hypothesis. By defining a loss function, we tell the
system what to do (i.e.\ the task to perform) or what to predict 
under each situation. Training the system with this generative 
process then encourages the system to adopt the different 
generators as potential hypotheses.

\subsubsection{Sequences}

Let $\mathcal{X}$ be a finite alphabet of observations. The set of
(finite) strings over ${\cal X}$ is written as $\mathcal{X}^{\ast}$,
which includes the empty string $\epsilon$. $\mathcal{X}^{\infty}$
denotes the set of one-way infinite sequences over $\mathcal{X}$.
For concreteness, we assume that $\mathcal{X}:=[N]:=\{1,2,\ldots,N\}$,
where $N$ could be very large. For strings, subindices correspond
to (time) indices as in~$x_{t}$, and we use obvious shorthands for
substrings such as $x_{t:t+k}:=x_{t}x_{t+1}\ldots x_{t+k}$, and $x_{<t}:=x_{1:t-1}$.
If the length $T$ is implicit from the context, then we also write
strings of length $T$ simply as $\tau$, from the word \emph{trajectory}.

\subsubsection{Generators/Hypotheses}

The domain of possible (stochastic) generators will be modeled using
a \emph{set of generators}, formalized as a class ${\cal P}$ of distributions
over infinite sequences in $\mathcal{X}^{\infty}$. These will become, after training, the \emph{set of hypotheses} of the system, and henceforth
we will use the terms ``generator'' and ``hypotheses'' interchangeably.
Specifically, we demand that
for each distribution $P\in{\cal P}$ over strings, the probability $P(x_{t}|x_{<t})$ of
any next symbol $x_{t}$ given any past $x_{<t}$ is specified.%
\footnote{This requirement avoids the technical subtleties 
associated to conditioning on pasts having probability zero.}
The probability of an observation string $x_{\leq t}$ is then equal
to the product $P(x_{\leq t})=\prod_{k=1}^{t}P(x_{k}|x_{<k})$. 
Defining the conditionals also uniquely determines the
distribution over infinite sequences.%
\footnote{More precisely, a collection of consistent distributions
may be defined over each of the spaces $\mathcal{X}^t$ for 
$t \in \mathbb{N}$ via the conditionals specified above. 
\emph{Kolmogorov's extension theorem} then states that
there exists a distribution $P$ over infinite sequences 
in $\mathcal{X}^{\infty}$ (with respect to the sigma-algebra 
generated by cylinder sets $\Gamma_{w}\subset\mathcal{X}^{\infty}$, 
where $\Gamma_{w}$ denotes the set containing all one-way 
infinite sequences having a common prefix $w\in\mathcal{X}^{\ast}$) 
that is consistent with the distributions over finite sequences 
defined via conditionals.}

We will also index the distributions in $\mathcal{P}$ by a
countable parameter set $\Theta$, so that each member is a 
distribution $P_{\theta}\in\mathcal{P}$,
where $\theta\in\Theta$.%
\footnote{This choice of the cardinality of the
set of parameters is for simplifying our mathematical
exposition. In practice, the extension to uncountable parameter sets
is straightforward (see the Dirichlet example below)} 
To use a notation that fits neatly the Bayesian
interpretation, we will write $P(x_{t}|\theta,x_{<t})$ rather than
$P_{\theta}(x_{t}|x_{<t})$; that is, where the parameter $\theta\in\Theta$
is interpreted as a conditional.

Finally we place prior probabilities $P(\theta)$ over the members
in $\Theta$. This will play the role of our measure of simplicity
(or inductive bias) of a hypothesis, where a simpler one possesses
more prior mass.%
\footnote{Measuring ``simplicity'' in this way is justified
by the implied description length of the hypothesis under an optimal
code, namely $-\log_2 P(\theta)$ bits.}

\begin{example}
\label{exa:hypothesis-dice-roll}(Dice roll prediction) In a dice
roll prediction problem, the set of hypotheses is given by a collection
of $N$-sided ``dice'' that generate i.i.d.~rolls according to the
categorical distribution, that is (with a slight abuse of notation)
$x\sim P(x=i|\theta)=\theta_{i}$,
where $\theta_{i}$ is the $i$-th element of the probability vector~$\theta\in\Delta([N])$.
If there are~$|\Theta|$ such dice, then a possible prior distribution
is the uniform $P(\theta)=\tfrac{1}{|\Theta|}$. 

Indeed, this example can be generalized to the set of all probability
vectors in the simplex~$\Delta([N])$, with a uniform prior density.
In this case, the whole process (i.e.~first sampling the parameter~$\theta$
and then sampling the observation sequence $x_{1},x_{2},\ldots$)
is known as a \emph{Dirichlet-Categorical process}. 
\end{example}

\begin{example}
(Optional: Algorithmic sequence prediction) In the introductory example we used
sequences that are generated according to patterns. These patterns
can be formalized as algorithms.%
\footnote{In fact, Example~\ref{exa:hypothesis-dice-roll} is a special case
of a collection of algorithms.} 
\emph{Algorithmic information theory} allows us to formalize a well-known,
very large hypothesis class \citep{vitanyi1997introduction}. Take
a universal Turing machine~$U$ that takes a binary-encoded program
$\theta\in\{0,1\}^{\ast}$ and produces a binary sequence $U(\theta)=x_{1}x_{2}\ldots$
as a result.%
\footnote{We assume that the universal Turing machine is prefix-free, and that
it outputs trailing zeros after reaching a halting state.}
Then we can generate a random binary sequence by providing the universal
Turing machine $U$ with fair coin flips, and then running the machine
on this input.

In this case, each hypothesis is a (degenerate) distribution over
sequences, where $P(x|\theta)=1$ if $x$ is a prefix of $U(\theta)$
and zero otherwise.
The prior distribution over binary programs is
$P(\theta)=2^{-l(\theta)}$, where $l(\theta)$ is the length of the
program~$\theta$.%
\footnote{Strictly speaking, this distribution might be
un-normalized, and we refer the reader to \citet{vitanyi1997introduction}
for a detailed technical discussion.}   
The resulting distribution over sequences is known
as the \emph{algorithmic prior} and \emph{Solomonoff's prior}.
\end{example}

\subsubsection{Strategies}

The system uses a strategy to solve a task. In general, strategies 
can implement predictions (over observations) and/or policies 
(distributions over actions). In the prediction case, we formally define 
a \emph{strategy} as a distribution $\pi$ over strings in ${\cal X}^{\infty}$.
This is probability distribution that characterizes the system's
outputs, and it should not be confused with the generators.
Then, $\pi(x_{t}|x_{<t})$ will denote a 
prediction over the next symbol $x_{t}$ given the past $x_{<t}$.%
\footnote{Throughpout the paper, for any distribution $\pi$ 
over ${\cal X}^\infty$, we define $\pi(x_{\le t})$ as the 
marginal distribution over the first $t$ symbols, i.e.\ %
$\pi(x_{\le t})= \pi(\{x' \in {\cal X}^\infty: x'_{\le t} = x_{\le t}\})$. 
Then, $\pi(x_{t}|x_{<t})$ is defined as 
$\pi(x_{t}|x_{<t})=\frac{\pi(x_{\le t})}{\pi(x_{<t})}$.}
The set of candidate strategies available to the agent is denoted as $\Pi$.

\subsubsection{Losses}

We consider tasks that can be formalized as the minimization of a
loss function. A \emph{loss function} is a function $\ell$ that maps
a strategy $\pi\in\Pi$ and a trajectory $\tau\in{\cal X}^{\ast}$
into a real-valued cost $\ell(\pi;\tau)\in\mathbb{R}$. Intuitively,
this captures the loss of using the strategy~$\pi$ under the \emph{partial observability}
of the parameter~$\theta$ when the trajectory is~$\tau$. For instance,
in sequential predictions, a typical choice for the loss function 
is the \emph{log-loss} $\ell(\pi, \tau) = -\log \pi(\tau)$, also known as the \emph{compression loss}; 
in sequential decision-making problems, one typically chooses 
a negative utility, such as the negative (discounted) 
cumulative sum of rewards.

\subsubsection{Goal}
The aim of the system is to minimize the \emph{expected loss}
\begin{equation}
\mathbf{E}[\ell]
 = \sum_{\theta} P(\theta) \Bigl[\sum_{\tau}P(\tau|\theta)\ell(\pi;\tau)\Bigr],\label{eq:expected-loss}
\end{equation}
with respect to the strategy $\pi$. That is, the objective is
the expected loss of a trajectory $\tau$, but generated by a \emph{latent} 
hypothesis~$\theta$ (i.e.\ the ground truth) randomly chosen according
to the desired inductive bias~$P(\theta)$.
This is the standard objective in Bayesian decision-theory in which the 
system has to choose in the face of (known) uncertainty
\citep{savage1972foundations}. Notice that this setup 
assumes the \emph{realizable} case, that is, the case 
in which the true generative
distribution is a member of the class of hypotheses
pondered by the system.

\subsection{Universality: Bayesian answer}

We start by showing how to solve the prediction problem from
a purely Bayesian point of view. This brief digression is
necessary in order to fully understand the statistical properties 
of the samples generated during meta-learning.
It is worth pointing out that the Bayesian solution is typically 
viewed as following directly from an interpretation of probabilities as 
degrees of belief rather than from an optimization problem \citep{jaynes2003probability}. This shouldn't be a distraction
however, as we will later express the Bayesian solution 
in terms of the compression loss in order to establish 
the connection to meta-learning.

The classical Bayesian approach
consists in using a predictor $\pi^\ast$ given by the \emph{mixture distribution}
\begin{equation}
\pi^\ast(\tau)=P(\tau)=\sum_{\theta}P(\theta)P(\tau|\theta)\label{eq:mixture}
\end{equation}
in which the probability of given trajectory~$\tau$ is given by
combining the probabilities predicted by the individual hypotheses
weighted by their prior probabilities. The prior probabilities are 
the initial inductive biases.

The mixture distribution automatically implements an \emph{adaptive}
prediction strategy, and not a static average predictor as a naive
reading of its definition might suggest. This is seen as follows.
The Bayes rule and \eqref{eq:mixture} easily imply that given past data~$x_{<t}$, we can predict the next observation~$x_{t}$
using the \emph{posterior predictive distribution} obtained by conditioning
$\pi^\ast$ on the past:
\begin{equation}
\pi^\ast(x_{t}|x_{<t})=P(x_{t}|x_{<t})=\sum_{\theta}P(\theta|x_{<t})P(x_{t}|\theta,x_{<t}).\label{eq:posterior-predictive}
\end{equation}
As (\ref{eq:posterior-predictive}) shows, the prediction of~$x_{t}$
is given by the average prediction made by the different hypotheses,
but weighted by the \emph{posterior probabilities} $P(\theta|x_{<t})$,
i.e.~the weights updated by the data.

These predictions converge to the true distribution with 
more data;%
\footnote{Notice however that in general 
the posterior probabilities $\pi(\theta|x_{<t})$ do not converge
unless we impose stricter conditions.}
that is, with probability one,
\begin{equation}
\label{eq:mix-consistent}
  \bigl( \pi^\ast(x_t|x_{<t}) - P(x_t|\theta^\ast,x_{<t}) \bigr)
  \rightarrow 0 
\end{equation}
where $\theta^\ast$ is the latent parameter of the true 
generator \citep[Theorem 3.19]{hutter2004universal} (In contrast,
notice that the posterior distribution $P(\theta|x_{<t})$ does 
\emph{not} converge in general).
From the standpoint of \emph{lossless compression}, 
Bayesian prediction is the optimal strategy.
The rate of convergence in \eqref{eq:mix-consistent}, or more precisely, 
the convergence rate of the average excess compression loss 
(a.k.a. regret) can easily be established (see, e.g., 
\citet{CeLu06}): for any $\theta^*$, $t$ and sequence $x_{\le t}$,
\begin{align*}
-\frac{1}{t}\log \pi^\ast(x_{\le t}) + \frac{1}{t} \log P(x_{\le t}|\theta^*) \le \frac{1}{t}\log P(\theta^*).
\end{align*}
That is, depending on the task-prior $P(\theta^*)$, 
the average excess compression loss converges to zero at an $O(1/t)$ rate.

To prepare the ground for the next section on meta-learning, 
we will characterize the construction of the mixture distribution 
in terms of a solution to an optimization problem. For this, 
we use a central result from information theory. 
As suggested above, we can use 
the \emph{compression loss}
to regress the statistics of the distribution over trajectories.
Formally, we choose $\ell(\pi;\tau)=-\log\pi(\tau)$.
Then, we optimize~(\ref{eq:expected-loss})
\begin{align}
\pi^{\ast} & =\arg\min_{\pi}\Bigl\{\sum_{\theta}P(\theta)\Bigl[\sum_{\tau}P(\tau|\theta)\ell(\pi;\tau)\Bigr]\Bigr\}\label{eq:expected-prediction-loss}\\
 & =\arg\min_{\pi}\Bigl\{-\sum_{\tau}P(\tau)\log\pi(\tau)\Bigr\}.\nonumber 
\end{align}
where~$P(\tau)=\sum_{\theta}P(\theta)P(\tau|\theta)$. The resulting
expected prediction loss is the \emph{cross-entropy} of $\pi(\tau)$
from the marginal $P(\tau)$, which implies the minimizer $\pi^{\ast}(\tau)=P(\tau)$;
that is, precisely the desired Bayesian mixture distribution having
the optimal prediction properties \citep{dawid1999prequential}.

\subsection{Universality: meta-learning answer}

\begin{figure}[th]
\begin{centering}
\includegraphics{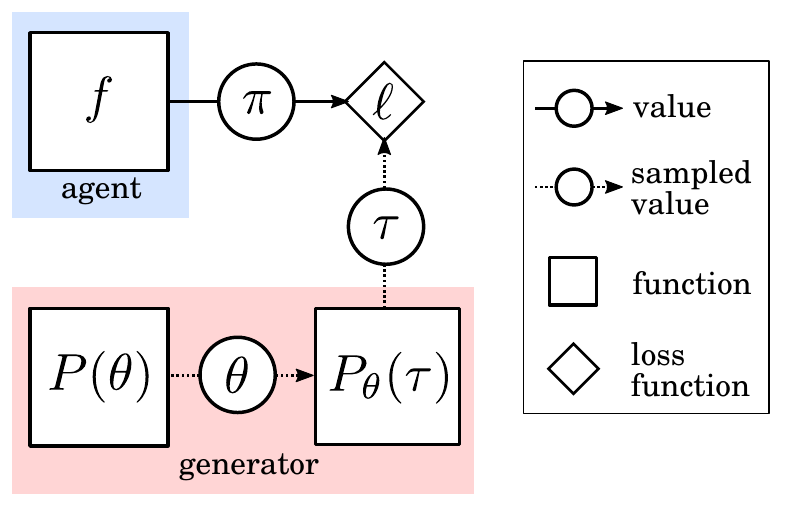}\caption{Basic computation graph for meta-learning a trajectory predictor. The
loss function depends only on the trajectory~$\tau$, not on the
parameter~$\theta$. Thus, the strategy~$\pi$ must marginalize
over the latent parameter~$\theta$.\label{fig:cg-basic}}
\par\end{centering}
\end{figure}

In a nutshell, meta-learning consists of finding a minimizer for a Monte-Carlo 
approximation of the expected loss~(\ref{eq:expected-loss}).
More precisely, consider a learning architecture (say, a deep learning architecture
trained by gradient descent) that implements a function class~${\cal F}$,
where each member $f\in{\cal F}$ maps a trajectory $\tau\in{\cal X}^{\ast}$
into a probability $f(\tau)=\pi(\tau)$ of the trajectory. Then, we
can approximate the expected loss~(\ref{eq:expected-loss}) as
\begin{equation}\label{eq:mc-loss}
\mathbf{E}[\ell]\approx\frac{1}{N}\sum_{n=1}^N\ell(f;\tau^{(n)}),
\end{equation}
where $\{\tau^{(n)}\}_{n}$ are $N$ i.i.d.\ samples drawn
from randomly chosen generators as 
\begin{align*}
\theta^{(n)} & \sim P(\theta)\\
\tau^{(n)} & \sim P(\tau|\theta^{(n)}).
\end{align*}
The goal of meta-learning is to find a function $f^\ast \in \cal{F}$ that minimizes
\eqref{eq:mc-loss}. Since the loss depends on the sampled trajectories $\tau^{(n)}$
but not on the parameters~$\theta^{(n)}$, the Monte-Carlo objective 
\emph{implicitly marginalizes} over the generators.%
\footnote{The technique of sampling 
the parameters rather than marginalizing over them explicitly was also called
\emph{root-sampling} in \citet{silver2010monte}.}
The computation graph is shown in Figure~\ref{fig:cg-basic}. 

We assume that optimizing the
Monte-Carlo estimate~(\ref{eq:mc-loss}) will give a minimizer~$f^{\ast}$
with the property
\[
f^{\ast}(\tau)\approx\pi^{\ast}(\tau)
\]
for the most probable~$\tau$.%
\footnote{We will avoid discussions 
of the approximation quality here. It suffices
to say that this depends on the learning model used; in particular,
on whether the optimum is realizable, and on the smoothness properties
of the model near the optimum.} 
While this gives us a method for modeling
the probabilities of trajectories, for sequential predictions we need
to incorporate some additional structure into the regression problem.

To do sequential predictions, i.e.~implementing $\pi^{\ast}(x_{t}|x_{<t})$,
the above optimization does not suffice; instead, we have to impose
the correct functional interface constraints onto the regression problem
in order to get a system that can map histories into predictions.
This is done by setting up the target loss so that the solution implements
the required function interface. Specifically, we seek a function
$f\in{\cal F}$ that maps histories $x_{<t}\in{\cal X}^{\ast}$ into
predictions $f(x_{<t})=\pi^{\ast}(x_{t}|x_{<t})\in\Delta({\cal X})$
of the next observation~$x_{t}\in{\cal X}$, thereby also respecting
the causal structure of the task. If we use a memory-based architecture,
such as a recurrent neural network 
\citep{elman1990finding, jordan1997serial, hochreiter2001learning, kolen2001field, graves2012supervised}, then the function $f$
\emph{conforms to the interface}
\begin{equation}
(\pi_t,m_t)=f(x_{t-1},m_{t-1}),\label{eq:prediction-interface}
\end{equation}
where $\pi_t \in\Delta({\cal X})$ is the current prediction vector,
$m_t$ and $m_{t-1}\in{\cal M}$ are the preceding and current memory states
respectively, and $x_{t-1}\in{\cal X}$ is the preceding observation. Furthermore, \emph{we fix the initial state}. This is a sufficient
condition for the system to make predictions that marginalize over
the hypotheses. Obviously, since $m_t$ must
remember all the necessary past information, the set of possible memory
states~${\cal M}$ needs to be sufficiently large for it to possess
the capacity for encoding the sufficient statistics of the pasts in~${\cal X}^{\ast}$.
See discussion below in Section~\ref{subsec:anatomy}. 

\begin{figure}[htb]
\begin{centering}
\includegraphics{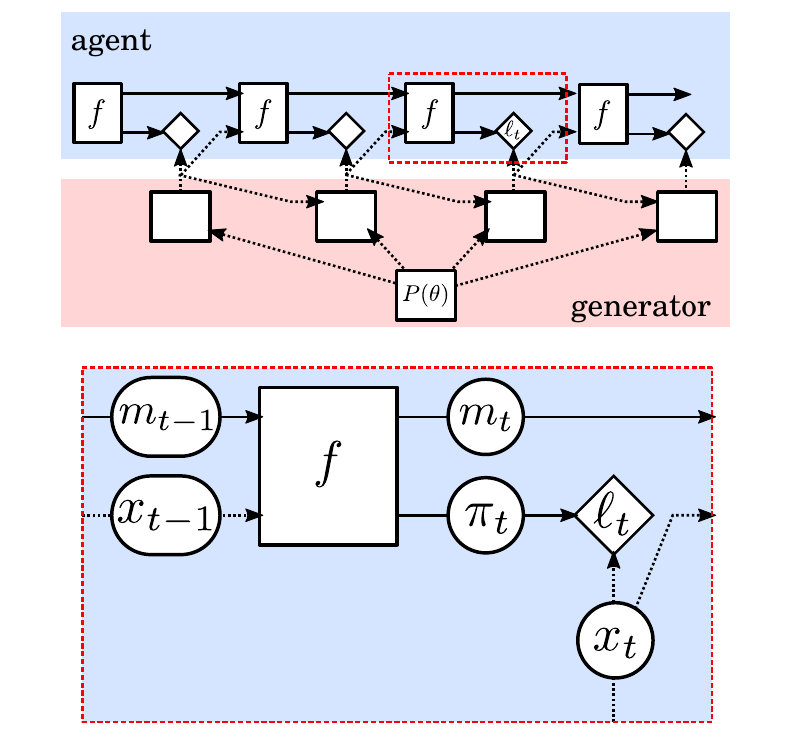}
\par\end{centering}
\caption{Computation graph for meta-learning a sequential 
prediction strategy. The agent function~$f$ generates a 
prediction~$\pi_{t}$ of the observation~$x_{t}$ based 
on the past, captured in the last observation~$x_{t-1}$
and the state~$m_{t-1}$. The top diagram illustrates 
the computation graph for a whole sequence (of length $T=4$), 
while the lower diagram shows a detailed view of a single step
computation.\label{fig:cg-seq-pred}}
\end{figure}

We need instantaneous predictions. The associated \emph{instantaneous
loss} function for this interface \eqref{eq:prediction-interface} is then
\begin{equation}
\ell_{t}=\ell(\pi_{t};x_{t})=-\log\pi_{t}(x_{t}),\label{eq:inst-log-loss}
\end{equation}
so that the Monte-Carlo approximation of the expected loss~(\ref{eq:mc-loss})
becomes
\begin{equation}
\frac{1}{N}\sum_{n=1}^N\ell(f;\tau^{(n)})=\frac{1}{N}\sum_{n=1}^N\sum_{t}^{T}\ell_{t}^{(n)},\label{eq:mc-sequential-loss}
\end{equation}
where $\ell^{(n)}_t$ are Monte-Carlo samples of the instantaneous log-loss. 
The computation graph is shown in Figure~\ref{fig:cg-seq-pred} and
the pseudo-code is listed in Algorithm~\ref{alg:prediction}.%
\footnote{Note that one could separate learning and sampling 
in Algorithm~\ref{alg:prediction} by first sampling $N$ models 
and corresponding observation sequences, and then feed the 
corresponding observation in the "observe" step of the algorithm. 
In later sections, however, the prediction of the agent will 
affect how the observations evolve; we chose to present sampling 
and prediction/learning in this interleaved way to have a more 
unified presentation style with Algorithms~\ref{alg:ts} 
and~\ref{alg:bayes-opt}, given in the next section.}
Arguably the most important result in the realizable setting is that,%
\footnote{Realizable is when the minimizer $f^{\ast}$ 
of the Monte-Carlo loss is in ${\cal F}$.}
if $N$ is large, and the minimum of~(\ref{eq:mc-sequential-loss})
is attained, then the minimizer~$f^{\ast}\in{\cal F}$ implements
a function
\[
f^{\ast}(x_{t-1},m_{t-1})=\bigl(\pi_{t},m_{t}\bigr)
\]
where, crucially, the instantaneous prediction~$\pi_{t}$ is 
\begin{equation}
\boxed{\pi_{t}(x_{t})\approx\pi^{\ast}(x_{t}|x_{<t})=\sum_{\theta}^{\vphantom{\theta}}P(\theta|x_{<t})P(x_{t}|\theta,x_{<t}),}\label{eq:amortized-prediction}
\end{equation}
i.e.~the optimal sequential prediction in which the Bayesian update
is automatically \emph{amortized} (or pre-computed) \citep{ritchie2016deep}.
In practice,
this means that we can use $f^{\ast}$ without updating the parameters
of the model (say, changing the weights of the neural network) to
predict the future from the past, as the prediction algorithm is automatically
implemented by the recursive function~$f^{\ast}$ with the help of
the memory state~$m_{t}$. The next section discusses how this is
implemented in the solution.

\begin{algorithm}[htb]
\DontPrintSemicolon
\SetCommentSty{emph}
\SetKwComment{Comment}{(}{)}
\KwData{Prior $P(\theta)$, generators $P(x_t|\theta, x_{<t})$, and
initial predictor $\bar{f}$, memory state $\bar{m}$, and observation $\bar{x}$.}
\KwResult{Meta-learned predictor $f$.}
  $f \leftarrow \bar{f}$ \Comment*{initialize function}
  \While{$f$ not converged}{
    $L \leftarrow 0$ \Comment*{reset loss}
    \For(\Comment*[f]{rollout batch}){$n = 1, 2, \ldots, N$}{
      $(x_0, m_0) \leftarrow (\bar{x}, \bar{m})$ \Comment*{reset memory state}
      $\theta \sim P(\theta)$ \Comment*{sample parameter}
      \For(\Comment*[f]{perform rollout}){$t = 1, 2, \ldots, T$}{
        $(\pi_t, m_t) \leftarrow f(x_{t-1}, m_{t-1})$
          \Comment*{predict}
        $x_t \sim P(x_t|\theta, x_{<t})$
          \Comment*{observe}
        $\ell_t = -\log \pi_t(x_t)$
          \Comment*{instantaneous loss}
        $L \leftarrow L + \ell_t$
          \Comment*{accumulate total loss}
      } 
    } 
    $f \leftarrow \mathrm{MinimizationStep}(f, L)$
      \Comment*{do an update step}
  }
  \Return{f}
\caption{Meta-learning a prediction strategy}\label{alg:prediction}
\end{algorithm}

\subsection{Anatomy of meta-learned agents\label{subsec:anatomy}}

\begin{figure}
\begin{centering}
\includegraphics[width=1\columnwidth]{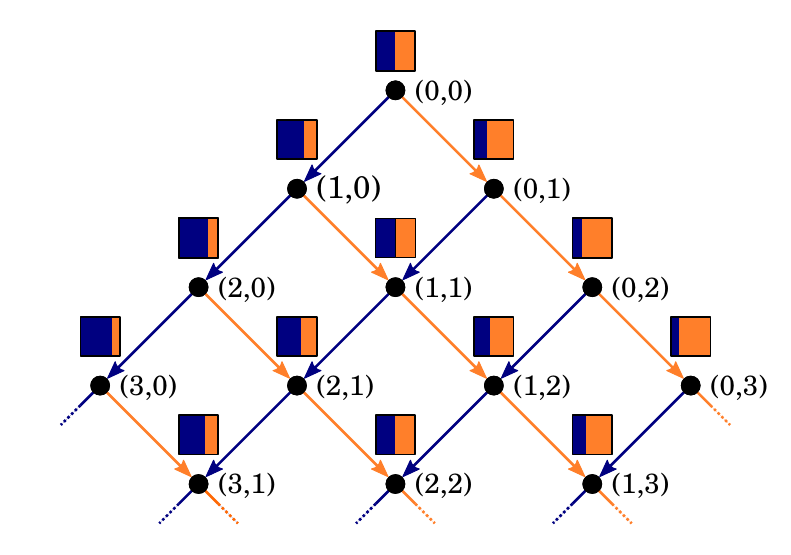}
\par\end{centering}
\caption{Minimal state machine for a predictor of coin tosses with a fixed,
unknown bias. The hypothesis class can be modeled as a 2-sided coin
(see Example~\ref{exa:hypothesis-dice-roll}). Dark and light state
transitions correspond to observing the outcomes `Head' and `Tail'
respectively, and the states are annotated with $(n_\text{H},n_\text{T})$, the number of times Head and Tail have been observed. The predictions made from each state are shown in the
stacked bar charts: the probability of Head is
$P(x_t=\text{H}|x_{<t})=\frac{n_\text{H}+1}{t+2}$ (which is
how these predictions are implemented in a computer program).
Note how different observations sequences can
lead to the same state (e.g.~HT and TH).\label{fig:minimal-state-machine}}
\end{figure}

\begin{figure}
\begin{centering}
\includegraphics[width=1\columnwidth]{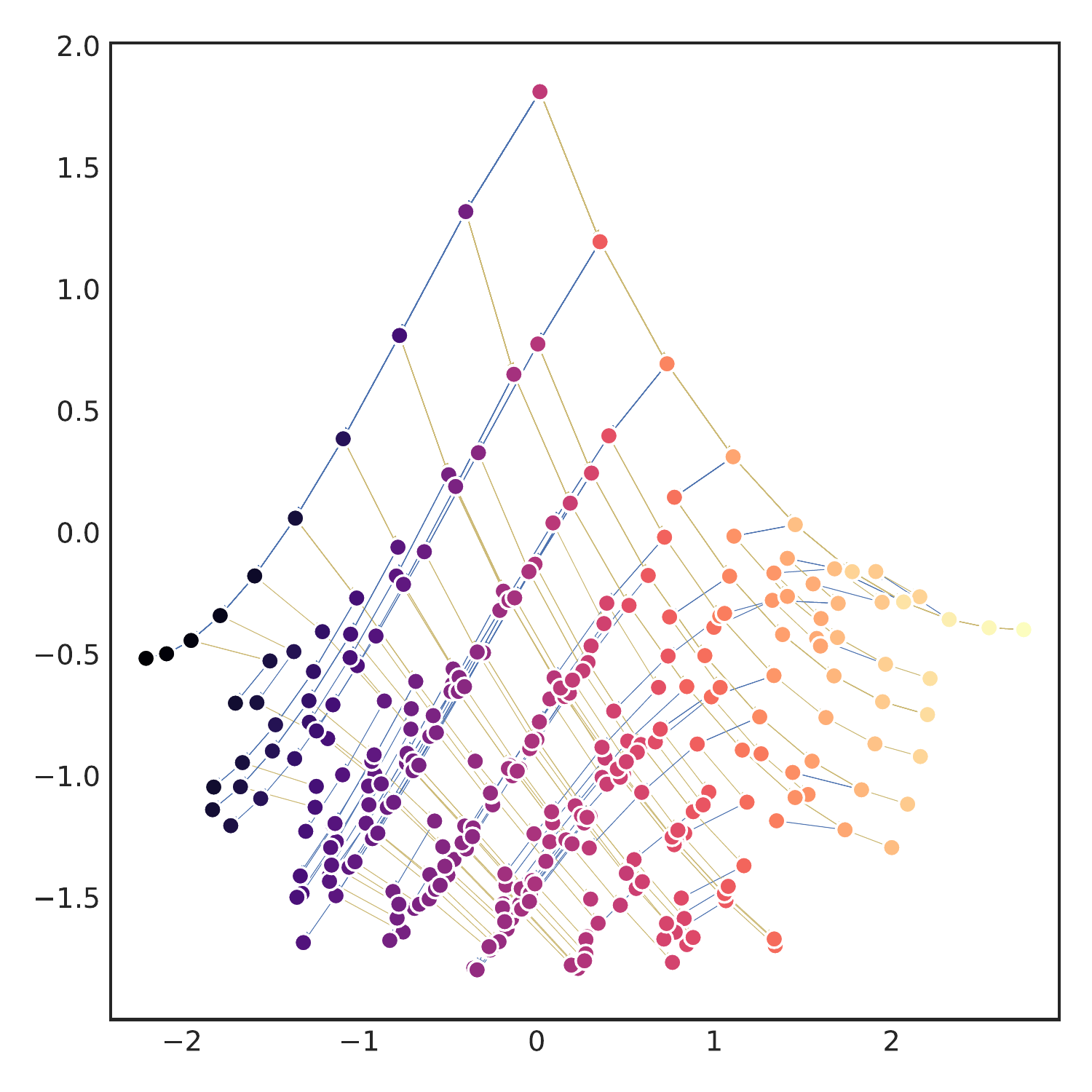}
\par\end{centering}
\caption{Meta-learned state machine for a predictor of coin tosses.
The figure shows the memory dynamics of a standard memory-based predictor 
projected onto the first two eigenvectors. Notice the
striking similarity with Figure~\ref{fig:minimal-state-machine}.
The predictor consists of 20 LSTM cells with softmax predictions,
which was trained using Algorithm~\ref{alg:prediction}
on 1000 batches of 100 rollouts,
where rollouts were of length 10. For training, we 
used the Adam optimization algorithm \citep{kingma2014adam}.
\label{fig:metalearned-state-machine}}
\end{figure}

How does the function $f^{\ast}\in{\cal F}$ implement the sequential
strategy with amortized prediction (\ref{eq:amortized-prediction})?
In which sense does this strategy rely on an internal model?
First, let us review the conditions:
\begin{enumerate}
\item \emph{Choice of loss function:} The choice of the loss function specifies
what solving the task means, i.e.~what we want the agent to do as
a function of the data. 
\item \emph{Functional interface:} Since the agent is ultimately implemented
as a function, the choice of the interface (e.g. mapping entire trajectories
into probabilities versus mapping pasts into predictions) is crucial.
Obviously this choice is implicit in the practice of any regression-based
machine learning technique. Nevertheless, we point this out because
it is especially important in sequential problems, as it determines
how the agent is \emph{situated} within the task, that is, what informational
(and ultimately causal) constraints the agent is subject to for solving
a task.
\item \emph{Monte-Carlo marginalization:} Marginalizing analytically over
the generators is in general intractable, except for special cases
such as in some exponential families \citep{koopman1936distributions,abramovich2013statistical}.
Instead, meta-learning performs a Monte-Carlo marginalization. Furthermore,
in the sequential setting, the marginalization hides the identity
of the generator, thereby forcing the model to find a function that
uses the past experience to improve on the loss. This, in turn, leads
to the numerical approximation of amortized Bayesian estimators, as
long as the first state is fixed across all the samples.
\end{enumerate}
As a result, we obtain a function~$f^{\ast}$ that performs the following
operations:
\begin{enumerate}
\item \emph{New prediction:} $f^{\ast}$ takes the past input $x_{t-1}$
and memory state $m_{t-1}$ to produce a new prediction $\pi_{t}$
minimizing the instantaneous loss $\ell_{t}$.
\item \emph{New memory state:} In order to perform the new prediction,
$f^{\ast}$ combines the past input~$x_{t-1}$ and memory state~$m_{t-1}$
to produce a new memory state~$m_{t}$ that acts as a sufficient
statistic of the entire past~$x_{<t}$. That is, there exists a sufficient
statistic function $\phi$ extracting all the necessary information from the past to predict the future, i.e.
\[
\pi(x_{t}|x_{<t})=\pi\bigl(x_{t}\bigl|\phi(x_{<t})\bigr),
\]
and this function is related to $f^\ast$
via the equation $\phi(x_{<t})=f^\ast(x_{t-1},m_{t-1})$.
\end{enumerate}
That is, the recursive function~$f^{\ast}$ implements a \emph{state
machine} \citep{sipser2006introduction} (or transducer) in which 
the states correspond to memory states, the transitions
are the changes of the memory state caused by an input, 
and where the outputs are the predictions.
We can represent this state machine as a labeled directed graph.
Figure~\ref{fig:minimal-state-machine} shows a state machine
predicted by theory and Figure~\ref{fig:metalearned-state-machine} shows
a meta-learned state machine.

The state machine thus implements a dynamics driven by the input,
where the state captures the sufficient statistics of the past \citep{kolen2001field}.
However, it is important to note that the objective~(\ref{eq:mc-sequential-loss})
does not enforce the minimality of the state machine (at least not
without post-processing \citep{kolen1994fool}). Indeed, in practice this is often
not the case, implying that the states do not correspond
to the \emph{minimal} sufficient statistics and that optimization runs with
different initial conditions (e.g.~random seeds) can produce different
state machines. Furthermore, as with any machine learning
method, the accuracy of the transitions of input-state 
pairs $(x,m)$ that never occurred during training depend 
on the generalization ability of the function approximator. 
This is especially the case for input sequences
that are longer than the length $T$ of the trajectories seen during
training.

State machines are important because they reflect symmetry relations
due to their intimate relation to semigroups \citep{krohn1968algebraic}.
If a node in the graph has two or more incoming arrows, then there
exist two past observation strings~$\tau^{(1)}$ and $\tau^{(2)}$,
not necessarily of the same length, such that
\[
\pi\bigl(\cdot\bigl|\phi(\tau^{(1)})\bigr)=\pi\bigl(\cdot\bigl|\phi(\tau^{(2)})\bigr),
\]
that is, they map onto the same sufficient statistics \citep{diaconis1988sufficiency},
and hence, all the trajectories that emanate from those states are
jointly amortized. Thus, analyzing the graph structure can reveal
the invariances of the data relative to the task. In particular, a
minimal state machine captures all the invariances. For instance,
an exchangeable stochastic sequence (in which the sequence is generated i.i.d. conditional on the hypothesis)
leads to a state machine with lattice structure as in Figure~\ref{fig:minimal-state-machine}.

\section{Sequential Decision-Making}\label{sec:seq-dm}

We now describe two ways of constructing interactive agents,
i.e.\ agents that exchange actions and observations with an external environment.
Many of the lessons learned in the sequential prediction case carry over.
The main additional difficulty is that, in the decision-making case, unlike in the prediction case,
 the optimal policy (which is
needed in order to generate the trajectories using the right distribution
for meta-learning) is not available. Hence, we need
to \emph{interleave two processes}: a meta-learning process that 
implicitly amortizes
the marginalization over the generators; and a policy improvement
process that anneals toward the optimal policy.

\subsection{Thompson sampling}

We can leverage the ideas from the sequential prediction case to create
an adaptive agent that acts according to probability matching\textemdash and
more specifically, \emph{Thompson sampling} \citep{thompson1933likelihood}%
\textemdash to address the exploration-exploitation problem 
\citep{sutton1998reinforcement}.
For this, we need generators that not only produce observations, but
also optimal actions provided by experts, which are then 
used as teaching signals.

In particular, provided we know the expert policies,
Thompson sampling translates the reinforcement learning
problem into an inference problem. Hence, meta-training 
a Thompson sampler is akin to meta-training a sequential
predictor, with the crucial difference that we want
our system to predict expert actions rather than 
observations. Due to this, Thompson samplers are optimal 
in the compression sense (i.e.\ using the log-loss), 
but not in the Bayes-optimal sense (see next subsection).

Formally, this time we consider distributions $P_{\theta}\in{\cal P}$
over interaction sequences, that is, strings in $({\cal A}\times{\cal O})^{\ast}$,
where ${\cal A}$ and ${\cal O}$ are discrete sets of actions and
observation respectively. We underline symbols to glue them together,
so $\underline{ao}_{t}:=(a_t, o_t)$. Then, a generator is a member
$P_{\theta}\in{\cal P}$ defining a distribution over strings 
\[
P(\underline{ao}_{\leq T}|\theta)=\prod_{t}^{T}P(a_{t}|\theta,\underline{ao}_{<t})P(o_{t}|\theta,\underline{ao}_{<t}a_{t})
\]
where the conditional probabilities
\[
P(o_{t}|\theta,\underline{ao}_{<t}a_{t})\quad\text{and}\quad P(a_{t}|\theta,\underline{ao}_{<t})
\]
are the probabilities of the next observation~$o_{t}$ and of the
next action~$a_{t}$ given the past, respectively. One can interpret
the $P(a_{t}|\theta,\underline{ao}_{<t})$ as the desired (or optimal)
policy provided by an expert when the observations follow 
the statistics $P(o_{t}|\theta,\underline{ao}_{<t}a_{t})$.
In addition, these probabilities must match the causal structure
of the interactions for our following derivation to be correct.%
\footnote{In practice, this is achieved by enforcing a 
particular factorization of the joint probability distribution 
over parameters and interactions into conditional probabilities 
that reflect the causal structure---see \citet{pearl2009causality}.}

Thompson sampling can be characterized as sampling the actions directly
from the posterior predictive \citep{ortega2010minimum}.
As in Bayesian prediction, consider the mixture distribution
\[
P(\underline{ao}_{\leq T})=\sum_{\theta}P(\underline{ao}_{\leq T}|\theta)P(\theta).
\]
Then we can generate actions by sampling them from the posterior predictive
\begin{equation}
a_{t+1} \sim P(a_{t+1}|\underline{\hat{a}o}_{\leq t}) 
= \sum_{\theta} P(a_{t+1}|\theta,\underline{ao}_{\leq t})
  P(\theta|\underline{\hat{a}o}_{\leq t})
  \label{eq:action-posterior-predictive}
\end{equation}
where the ``hat'' as in ``$\hat{a}$'' denotes a causal intervention and
where $P(\theta|\underline{\hat{a}o}_{\leq t})$ is recursively
given by%
\footnote{See \citet{pearl2009causality} for a thorough definition of 
causal interventions. Equations \eqref{eq:action-posterior-predictive} 
and \eqref{eq:causal-posterior} are non-trivial and beyond the scope of this report; 
we refer the reader to \citet{ortega2010minimum} for their derivation.
In particular, note that $P(a_{t+1}|\theta,\underline{ao}_{\leq t})$ in
\eqref{eq:action-posterior-predictive} does not have interventions.}
\begin{equation}
P(\theta|\underline{\hat{a}o}_{\leq t}) = 
\frac{P(o_t|\theta, \underline{ao}_{<t}) P(\theta|\underline{\hat{a}o}_{<t})}
     {\sum_{\theta'} P(o_t|\theta', \underline{ao}_{<t}) P(\theta'|\underline{\hat{a}o}_{<t})}
\label{eq:causal-posterior}
\end{equation}
In other words, we continuously condition on the past, treating actions 
as interventions and observations as normal (Bayesian) conditions. 
More precisely, unlike observations,
past actions were generated by the agent without knowledge of the 
underlying parameter (hence the $a_{t}$ and $\theta$ are 
independent conditional on the past experience), and
the causal intervention mathematically accounts for this fact.

Meta-learning a Thompson sampling agent follows a scheme analogous
to sequential prediction. We seek a strategy~$\pi^\ast$ that amortizes
the posterior predictive over actions~(\ref{eq:action-posterior-predictive}):
\[
\boxed{\pi^\ast(a_{t}|\underline{ao}_{<t})\approx P(a_{t}|\underline{\hat{a}o}_{<t})=\sum_{\theta}^{\vphantom{\theta}}P(a_{t}|\theta,\underline{ao}_{<t})P(\theta|\underline{\hat{a}o}_{<t}).}
\]
This strategy conforms to the functional interface
\begin{equation}
(\pi_t,m_t)=f(a_{t-1},o_{t-1},m_{t-1}),\label{eq:ts-interface}
\end{equation}
where $\pi_t\in\Delta({\cal A})$ is the current policy vector, $m_t$ and $m_{t-1}\in{\cal M}$ are the preceding and current
memory states respectively, and $\underline{ao}_{t-1}\in{\cal A}\times{\cal O}$ is the preceding interaction.

\begin{figure}[th]
\begin{centering}
\includegraphics{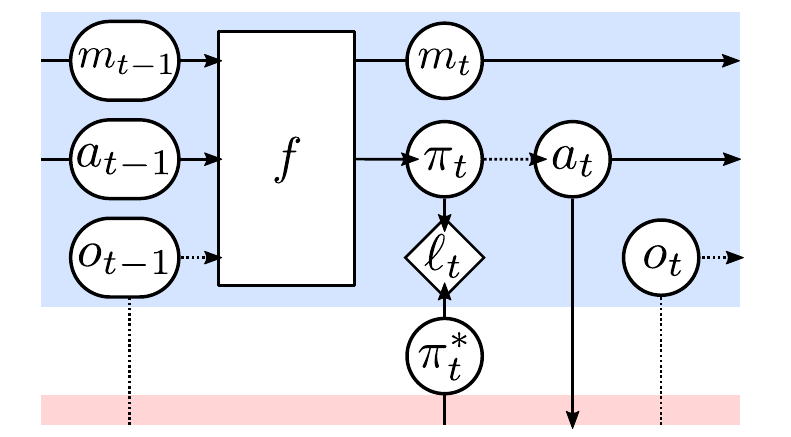}\caption{Detail of the computation graph for meta-learning a Thompson sampling agent. The actions $a_t$ are generated from the agent's policy $\pi_t$. The expert policy $\pi^\ast_t$ is only used for generating a loss signal $\ell_t$.\label{fig:cg-ts}}
\par\end{centering}
\end{figure}

Next we derive the loss function. As in Bayesian prediction, Thompson
sampling optimizes the compression of the interaction sequence characterized
by the expected log-loss. This is easiest written recursively in terms
of the instantaneous expected log-loss as 
\begin{equation}
\mathbb{E}\biggl[
-\sum_{\theta}P(\theta|\underline{\hat{a}o}_{<t})\Bigl[\sum_{a_{t}}P(a_{t}|\theta,\underline{ao}_{<t})\log\pi(a_{t}|\underline{ao}_{<t})\Bigr]
\biggr]
\label{eq:ts-log-loss}
\end{equation}
for each action $a_t$ given its past $\underline{ao}_{<t}$.%
\footnote{The expected log-loss for observations is omitted, 
as we only need to regress the policy here.}
For a Monte-Carlo approximation, we sample trajectories 
$\{\tau^{(n)}\}_{n}$ as
\begin{equation}
\left.\begin{aligned}\theta^{(n)} & \sim P(\theta)\\
a_{t}^{(n)} & \sim\pi(a_{t}|\underline{ao}_{<t}^{(n)})\quad\text{for all \ensuremath{t}}\\
o_{t}^{(n)} & \sim P(o_{t}|\theta^{(n)},\underline{ao}_{<t}^{(n)}a_{t}^{(n)})\quad\text{for all \ensuremath{t}}.
\end{aligned}
\right\} \label{eq:ts-sampling}
\end{equation}
In particular, note how actions are drawn from the agent's policy,
not from the generator. This ensures that the agent does not use 
any privileged information about the generator's identity, 
thus covering the support over
all the trajectories that the agent might explore. 
Then, we can choose the instantaneous loss function as the cross-entropy
\begin{equation}
\ell_{t}=-\sum_{a_{t}}P(a_{t}|\theta,\underline{ao}_{<t})\log\pi_{t}(a_{t})\label{eq:expected-policy-log-loss}
\end{equation}
evaluated on the sampled trajectories. This is known as (a variant of) \emph{policy
distillation} \citep{rusu2015policy}. Again, this only works if
we have access to the ground-truth optimal policy for each environment. 
The computation graph is shown in Figure~\ref{fig:cg-ts} and the pseudo-code is listed in Algorithm~\ref{alg:ts}

\begin{algorithm}[htb]
\DontPrintSemicolon
\SetCommentSty{emph}
\SetKwComment{Comment}{(}{)}
\KwData{Prior $P(\theta)$, generators $P(o_t|\theta, \underline{ao}_{<t}a_t)$, and policies $P(a_t|\theta, \underline{ao}_{<t})$; and
initial predictor $\bar{f}$, memory state $\bar{m}$, and interaction $\bar{a}, \bar{o}$.}
\KwResult{Meta-learned predictor $f$.}
  $f \leftarrow \bar{f}$ \Comment*{initialize function}
  \While{$f$ not converged}{
    $L \leftarrow 0$ \Comment*{reset loss}
    \For(\Comment*[f]{rollout batch}){$n = 1, 2, \ldots, N$}{
      $(a_0, o_0, m_0) \leftarrow (\bar{a}, \bar{o}, \bar{m})$ \Comment*{reset memory state}
      $\theta \sim P(\theta)$ \Comment*{sample parameter}
      \For(\Comment*[f]{perform rollout}){$t = 1, 2, \ldots, T$}{
        $(\pi_t, m_t) \leftarrow f(a_{t-1}, o_{t-1}, m_{t-1})$
          \Comment*{policy}
        $a_t \sim \pi_t(a_t)$
          \Comment*{act}
        $o_t \sim P(o_t|\theta, \underline{ao}_{<t}a_t)$
          \Comment*{observe}
        $\ell_t = -\sum_{a_t} P(a_t|\underline{ao}_{<t})\log \pi_t(a_t)$
          \Comment*{inst. loss}
        $L \leftarrow L + \ell_t$
          \Comment*{accumulate total loss}
      } 
    } 
    $f \leftarrow \mathrm{MinimizationStep}(f, L)$
      \Comment*{do an update step}
  }
  \Return{f}
\caption{Meta-learning a Thompson sampler}\label{alg:ts}
\end{algorithm}

Finally, we note that the above meta-training algorithm is 
designed for finding agents that implicitly update their posterior 
after each time step. However, this can lead to unstable
policies that change their behavior (i.e.\ expert policy) 
in each time step. Such inconsistencies can be addressed
by updating the posterior only after experiencing
longer interaction sequences---for instance, only
after an episode. We refer the reader to \citep{russo2018tutorial,
osband2016posterior, ouyang2017learning} for a detailed
discussion.

\subsection{Bayes-Optimality}

Bayes-optimal sequential decision-making is the decision strategy
that follows from the theory of \emph{subjective expected utility}
\citep{savage1972foundations} and the method of \emph{dynamic programming}
\citep{bellman1954theory}. Roughly, it consists in always picking
an action that maximizes the \emph{value}, that is, the expected sum
of future rewards under the best future actions. Methodologically,
it requires solving a stochastic partial difference equation modeling
the \emph{value} for given boundary conditions, where the latter typically
constrain the value to zero everywhere along the planning horizon
\citep{bertsekas2008neuro}. Due to this, learning a Bayes-optimal
policy is far more challenging than learning a Thompson sampling strategy.
Here we also depart from the log-loss, and use a reward function
instead to characterize the task goals.

In this case the generators are distributions $P_{\theta}\in{\cal P}$ 
over observation sequences conditioned on past interactions, 
where each member $P_{\theta}\in{\cal P}$ defines a conditional
distribution 
\[
P(o_{\leq T}|\theta,\underline{ao}_{<t}a_{t})=\prod_{t}^{T} P(o_{t}|\theta,\underline{ao}_{<t}a_{t}).
\]
However, unlike the Thompson sampling case, here we seek a global
optimal policy $P(a_{t}|\underline{ao}_{<t})$ which is indirectly
defined via a reward function as discussed later. This global policy 
will by construction solve the exploration-exploitation
problem in a Bayes-optimal way, although it is tailored 
specifically to the given task distribution.

Because the optimal policy is unknown, in practice during training
we only have access to trajectories~$\tau$ that are drawn from a
distribution~$P^{\pi}$ that results from the interactions between
the expected task and a custom policy $\pi$. $P^{\pi}$ is given by
\[
P^{\pi}(\underline{ao}_{\leq T})=\sum_{\theta}P(\theta)\prod_{t}^{T}\pi(a_{t}|\underline{ao}_{<t})P(o_{t}|\theta,\underline{ao}_{<t}a_{t}),
\]
that is, a distribution where actions are drawn from the agent's current
strategy~$\pi$ and where observations are drawn from a randomly
chosen generator as described in~(\ref{eq:ts-sampling}) from Thompson
sampling. 

As mentioned above, we specify the agent's objective with a reward function.
This is a global reward function $r$ that maps every interaction
$\underline{ao}_{t}\in {\cal A}\times{\cal O}$ and past $\underline{ao}_{<t}\in({\cal A}\times{\cal O})^{\ast}$
into a scalar value $r(\underline{ao}_{t}|\underline{ao}_{<t})\in\mathbb{R}$,
indicating the interaction's desirability. Furthermore, we define
the action-value function for a policy~$\pi$ as the expected sum
of rewards given a past,%
\footnote{For simplicity, we assume that rewards are undiscounted.} 
that is,
\begin{equation}
Q^{\pi}(a_{t}|\underline{ao}_{<t})=\mathbf{E}^{\pi}\Bigl[\sum_{k=t}^{T}r(\underline{ao}_{k}|\underline{ao}_{<k})\Bigl|\underline{ao}_{<t}a_{t}\Bigr]
\label{eq:q-value}
\end{equation}
where the expectation $\mathbf{E}^{\pi}$ denotes an expectation w.r.t.~the
distribution~$P^{\pi}$. Notice that this definition implicitly equates
the rewards after the horizon~$T$ with zero. An optimal policy
$P(a_{t}|\underline{ao}_{<t})$ is defined as any policy that
maximizes~\eqref{eq:q-value} for any past $\underline{ao}_{<t}$.
Note that this is a recursive definition that can be solved using 
dynamic programming.

Meta-learning a Bayes-optimal policy can be done in numerous ways:
here we settle on inferring the optimal policy via estimating the 
action-values for concreteness, but other methods (e.g.\ using policy
gradients) work as well.  We
seek amortizing the action-values using a vector~$q_{t}\in\mathbb{R}^{|{\cal A}|}$:
\[
\boxed{q_{t}(a_{t})\approx Q^{\pi}(a_{t}|\underline{ao}_{<t}).}
\]
The functional interface conforms to
\[
(q_t,m_t)=f(a_{t-1},o_{t-1},m_{t-1})
\]
where $q_t\in\mathbb{R}^{|{\cal A}|}$ is the current action-value vector used
for constructing the policy; $m_t$ and $m_{t-1}\in{\cal M}$ are the preceding
and current memory states respectively; and $\underline{ao}_{t-1}\in{\cal A}\times{\cal O}$
is the preceding interaction, which implicitly also provides the reward.%
\footnote{If the reward is not a function of actions and observations,
then it needs to be passed explicitly alongside the last 
interaction.}

As the instantaneous loss-function~$\ell_{t}$ for regressing the
action-values, we can for instance use the TD-error
\[
\ell_{t}=\Bigl\{\bigl(r_{t}+q_{t+1}(a_{t+1})\bigr)-q_{t}(a_{t})\Bigr\}^{2}.
\]
Crucially, this is \emph{only} a function of the value $q_{t}(a_{t})$
of the current action. The \emph{target value}~$\bigl(r_{t}+q_{t+1}(a_{t+1})\bigr)$,
given by the sum of the current reward $r_{t}=r(\underline{ao}_{t}|\underline{ao}_{<t})$
and the value~$q_{t+1}(a_{t+1})$ of the next action, is kept constant,
ensuring that the boundary conditions are propagated in the right
direction, namely backwards in time.

To regress the optimal policy, we use \emph{simulated annealing} \citep{kirkpatrick1983optimization}.
Specifically, we start from a random policy and then slowly crystallize
an optimal policy. To do so, actions are drawn as 
$a_{t}\sim\pi_{t}(a_{t})$ from a policy built
from the action-values using e.g.\ the softmax function
\[
\pi_{t}(a_{t})=\frac{\exp\{\beta q_{t}(a_{t})\}}{\sum_{a}\exp\{\beta q_{t}(a)\}},
\]
where the inverse temperature $\beta>0$ is a parameter controlling
the stochasticity of the policy:~$\beta\approx0$ yields a nearly
uniform policy and~$\beta\gg0$ a nearly deterministic one. During
meta-training, the inverse temperature is annealed (cooled), 
starting from~$\beta=0$ and ending in a large value for~$\beta$. 
This gives the model time to regress the action-values by 
sampling sub-optimal branches before
committing to a specific policy. Good cooling schedules are typically
determined empirically \citep{mitra1986convergence,nourani1998comparison}.
The pseudo-code is listed in Algorithm~\ref{alg:bayes-opt}.

\begin{algorithm}[htb]
\DontPrintSemicolon
\SetCommentSty{emph}
\SetKwComment{Comment}{(}{)}
\KwData{Prior $P(\theta)$, generators $P(o_t|\theta, \underline{ao}_{<t}a_t)$, and reward function $r(\underline{ao}_t|\theta, \underline{ao}_{<t})$; and
initial predictor $\bar{f}$, inverse temperature $\bar{\beta}$, memory state $\bar{m}$, and interaction $\bar{a}, \bar{o}$.}
\KwResult{Meta-learned predictor $f$.}
  $f \leftarrow \bar{f}$ \Comment*{initialize function}
  $\beta \leftarrow \bar{\beta}$ \Comment*{initialize inv. temp.}
  \While{$f$ not converged}{
    $L \leftarrow 0$ \Comment*{reset loss}
    \For(\Comment*[f]{rollout batch}){$n = 1, 2, \ldots, N$}{
      $(a_0, o_0, m_0) \leftarrow (a, o, m)$ \Comment*{reset memory state}
      $\theta \sim P(\theta)$ \Comment*{sample parameter} 
      \For(\Comment*[f]{perform rollout}){$t = 1, 2, \ldots, T$}{
        $(q_t, m_t) \leftarrow f(a_{t-1}, o_{t-1}, m_{t-1})$
          \Comment*{Q-values}
        $\pi_t(a_t) \leftarrow \frac{\exp(\beta q_t(a_t))}{\sum_a\exp(\beta q_t(a))}$
          \Comment*{policy}
        $a_t \sim \pi_t(a_t)$
          \Comment*{act}
        $o_t \sim P(o_t|\bar{\theta}, \underline{ao}_{<t}a_t)$
          \Comment*{observe}
        $r_t \leftarrow r(\underline{ao}_t|\theta, \underline{ao}_{<t}a_t)$
          \Comment*{reward}
        \If(\Comment*[f]{compute previous loss}){$t-1 \geq 1$}{
          $q_{\mathrm{target}} \leftarrow \mathrm{Constant}(r_{t-1} + q_t(a_t))$\;
          $\ell_{t-1} = \bigl( q_\mathrm{target} - q_{t-1}(a_{t-1}) \bigr)^2$\;
          $L \leftarrow L + \ell_{t-1}$
            \Comment*{accumulate total loss}
        }
      } 
      $q_{\mathrm{target}} \leftarrow \mathrm{Constant}(r_T)$
          \Comment*{last target}
      $\ell_T = \bigl( q_\mathrm{target} - q_T(a_T) \bigr)^2$
          \Comment*{last inst. loss}
      $L \leftarrow L + \ell_T$
          \Comment*{accumulate total loss}
    } 
    $f \leftarrow \mathrm{MinimizationStep}(f, L)$
      \Comment*{do an update step}
    $\beta \leftarrow \mathrm{CoolingSchedule}(\beta)$
      \Comment*{update inv. temp.}
  }
  \Return{f}
\caption{Meta-learning a Bayes-optimal policy}\label{alg:bayes-opt}
\end{algorithm}

\section{Discussion}\label{sec:discussion}

\subsection{Meta-learning and Bayesian statistics}

Meta-learning is intimately connected to Bayesian statistics \emph{regardless}
of the loss function, due to the statistics of the generated trajectories.
When regressing a sequential strategy using a Monte-Carlo estimation,
we sample trajectories $\{\tau^{(n)}\}_{n}$ as%
\footnote{For sequential-decision making problems, we assume that the policy
improvement steps have already converged. In this case, the actions
are drawn from the target distribution for meta-learning, and we can
safely ignore the distinction between actions and observations.}
\begin{align*}
\theta^{(n)} & \sim P(\theta)\\
x_{t}^{(n)} & \sim P(x_{t}|\theta^{(n)},x_{<t}^{(n)})\quad\text{for all \ensuremath{t}.}
\end{align*}
However, from the point of view of the system that
has already seen the past~$x_{<t}$, the transition~$x_{t}$ 
looks, on average, as if it were sampled with probability
\begin{equation}
P(x_{t}|x_{<t})=\sum_{\theta}P(x_{t}|\theta,x_{<t})P(\theta|x_{<t}),\label{eq:bs-mixture}
\end{equation}
that is, from the Bayesian posterior predictive distribution, which in turn induces the (implicit) update of the hypothesis
\begin{equation}
P(\theta|x_{\leq t})\propto P(\theta|x_{<t})P(x_{t}|\theta,x_{<t}).\label{eq:bs-update}
\end{equation}
Hence, (\ref{eq:bs-mixture}) and~(\ref{eq:bs-update}) together
show that the samples are implicitly filtered according to Bayes'
rule. It is precisely this statistical property that is harvested
through memory-based meta-learning. As shown in Section~\ref{sec:seq-pred}, 
meta-learning a Bayesian sequence predictor corresponds 
to directly regressing the statistics of the samples; 
due to this, it can be considered the most basic form 
of meta-learning. In contrast, the two sequential decision 
makers from Section~\ref{sec:seq-dm} do not regress the 
statistics directly but rather use their correspondence to 
a Bayesian filtration to build adaptive policies.

Conceptually, using a generative process to produce 
Bayes-filtered samples is an old idea. It is the rationale 
driving many practical implementations of Bayesian models 
\citep{bishop2016pattern}, and one of the key ideas in Monte-Carlo 
methods for Bayes-optimal planning \citep{silver2010monte, 
guez2012efficient}.

\subsection{Sample complexity of strategies}

Current model-free RL algorithms such as deep RL algorithms (e.g.\ DQN
\citep{mnih2013playing} or A3C \citep{mnih2016asynchronous}) are known
to be sample inefficient. The sample complexity can potentially
be improved by using a suitable model-based approach with strong
inductive biases. Full probabilistic model-based approaches (i.e.~those
that do not work with expectation models) can rely on hand-crafted
probabilistic models that possess little structure (e.g.\ Dirichlet
priors over state transitions) or have intractable (exact) posterior
distributions. This can make such approaches unwieldy in practice.

More commonly, traditional approaches have used expectation models
\citep{sutton2012dyna, schmidhuber1990making}, or deterministic
abstract models \citep{watter2015embed, predictron}. However, so far
there has been limited success scaling probabilistic
modeling to improve sample efficiency in the context
of deep reinforcement learning.

Meta-learning addresses this problem in a conceptually straightforward
manner. It automates the synthesis of near-optimal algorithms, by searching
in algorithm space (or automata space) in order to find a new reinforcement
learning algorithm that is tailored to a given class of tasks, exploiting the
structure and using the desired inductive biases. 
\emph{The meta-learned algorithms minimize the sample 
complexity at test time because they directly minimize 
the expected loss averaged over all generators.} 
In the examples we have seen, meta-learning finds sequential algorithms, 
which when deployed perform near-optimally, minimizing the sample complexity.
The flip side is that meta-learning can be very expensive at
meta-training time due to the
slow convergence of the Monte-Carlo approximation and the very large
amount of data required by current popular neural architectures 
during the meta-training phase. 

\subsection{Spontaneous meta-learning}

Meta-learning can also occur spontaneously in online regression when
the capacity of the agent is bounded and the data is produced by a
\emph{single generator}. Unfortunately, the downside is that we cannot
easily control \emph{what} will be meta-learned. In particular, spontaneous
meta-learning could lead to undesirable emergent properties, which is
considered an open research problem in AI safety \citep{ortega2018building}. 
To see how meta-learning happens, 
consider the sequential prediction case. All we need is to show how 
the conditions for meta-learning occur naturally; that is, by identifying the
Monte-Carlo samples and their latent parameters.

\begin{table*}[t]
\caption{Example segmentation of an input sequence in spontaneous meta-learning.\label{tab:segmentation}}
\centering{}%
\begin{tabular}{rcc|cc|ccc|ccc|c}
Input: & $x_{1}$ & $x_{2}$ & $x_{3}$ & $x_{4}$ & $x_{5}$ & $x_{6}$ & $x_{7}$ & $x_{8}$ & $x_{9}$ & $x_{10}$ & $\cdots$\tabularnewline
Agent State: & $m_{1}$ & $m^{\ast}$ & $m_{3}$ & $m^{\ast}$ & $m_{5}$ & $m_{6}$ & $m^{\ast}$ & $m_{8}$ & $m_{7}$ & $m^{\ast}$ & $\cdots$\tabularnewline
Generator State: & $z_{1}$ & $z_{2}$ & $z_{3}$ & $z_{4}$ & $z_{5}$ & $z_{6}$ & $z_{7}$ & $z_{8}$ & $z_{9}$ & $z_{10}$ & $\cdots$\tabularnewline
 &  &  &  &  &  &  &  &  &  &  & \tabularnewline
Trajectory: & \multicolumn{2}{c|}{(na)} & \multicolumn{2}{c|}{$\tau^{(1)}$} & \multicolumn{3}{c|}{$\tau^{(2)}$} & \multicolumn{3}{c|}{$\tau^{(3)}$} & $\cdots$\tabularnewline
Derived Input: & \multicolumn{2}{c|}{(na)} & $y_{1}^{(1)}$ & $y_{2}^{(1)}$ & $y_{1}^{(2)}$ & $y_{2}^{(2)}$ & $y_{3}^{(2)}$ & $y_{1}^{(3)}$ & $y_{2}^{(3)}$ & $y_{3}^{(3)}$ & $\cdots$\tabularnewline
Derived Parameter: & \multicolumn{2}{c|}{(na)} & \multicolumn{2}{c|}{$\theta^{(1)}=z_{2}$} & \multicolumn{3}{c|}{$\theta^{(2)}=z_{4}$} & \multicolumn{3}{c|}{$\theta^{(3)}=z_{7}$} & $\cdots$\tabularnewline
\end{tabular}
\end{table*}

\begin{figure}[th]
\begin{centering}
\includegraphics[width=\columnwidth]{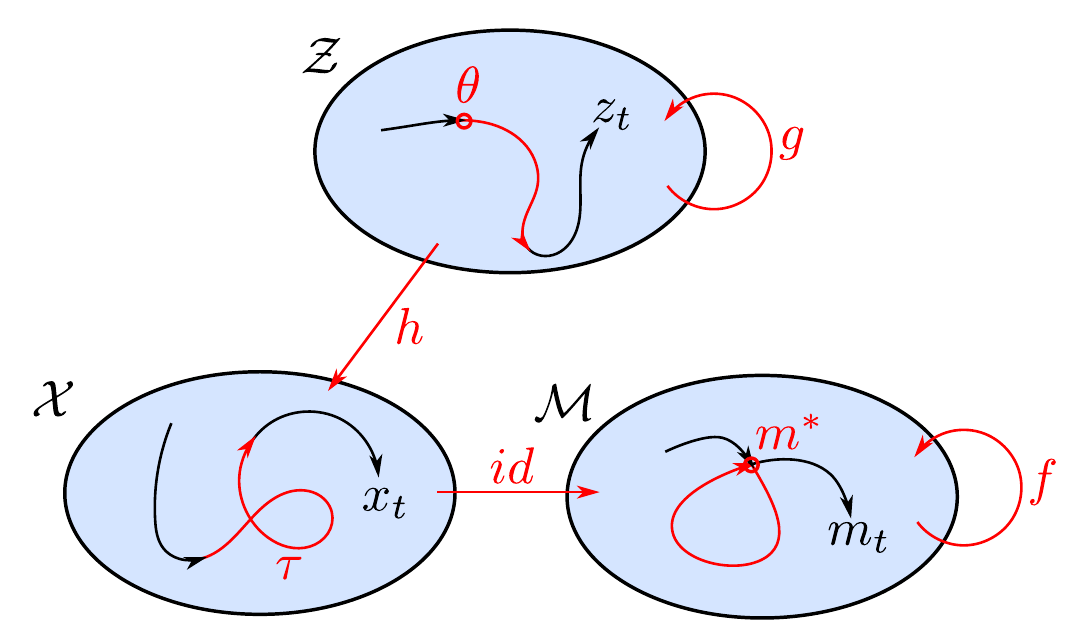}
\par\end{centering}
\caption{Dynamics of spontaneous meta-learning. The (pseudorandom) generator with states in $\cal Z$ produces observations in $\cal X$, which drive the agent's memory state-transitions in $\cal M$. A loop in the memory space with endpoints equal $m^\ast \in \cal M$ corresponds to a trajectory $\tau$ generated by a latent parameter~$\theta$.\label{fig:ml-online-dyn}}
\end{figure}

Let ${\cal M}$ and ${\cal Z}$ be the agent's and the generator's
set of internal states respectively, and let ${\cal X}$ be the set
of observations. Assume that the dynamics are deterministic 
(stochastic dynamics can be modeled using pseudo-random transitions)
and implemented by functions $f,g,h$, so that the sequence of observations~$x_{1},x_{2},\ldots\in{\cal X}$,
agent states~$m_{1},m_{2},\ldots\in{\cal M}$, and generator states~$z_{1},z_{2},\ldots\in{\cal Z}$
are given by
\begin{align*}
x_{t} & =h(z_{t}) & \text{} & \text{(observations)}\\
m_{t+1} & =f(x_{t},m_{t}) &  & \text{(agent states)}\\
z_{t+1} & =g(z_{t}), &  & \text{(generator states)}
\end{align*}
(with arbitrary initial values $z_{0}$, $m_{0}$) as illustrated
in Figure~\ref{fig:ml-online-dyn}. Furthermore, assume that~${\cal M}$
and~${\cal Z}$ are \emph{finite}.%
\footnote{This assumption can be relaxed to compact metric spaces using tools
from dynamical systems theory that are beyond the scope of this report.}
Then, for a sufficiently long sequence there must exist a state~$m^{\ast}\in{\cal M}$
that is visited infinitely often when given an infinite stream of observations.
We can use~$m^{\ast}$ to segment the
sequence of observations into (variable-length) trajectories $\tau^{(1)},\tau^{(2)},\ldots$
which we can identify as Monte-Carlo samples from a class of generators
as illustrated by the example in Table~\ref{tab:segmentation}. The
$n$-th trajectory $\tau^{(n)}$ is defined by the first substring
delimited by the $n$-th and $(n+1)$-th occurrence of $m^{\ast}$,
and define $y_{t}^{(n)}\in{\cal X}$ as the $t$-th observation of
the $n$-th trajectory, e.g.~$y_{1}^{(2)}=x_{5}$ because $\tau^{(2)}=x_{5}x_{6}x_{7}$.
Finally, the task parameter~$\theta^{(n)}$ of the $n$-th trajectory
is the state of the generator at the beginning of the sequence.

Given this identification, an online learning algorithm that updates~$f$
based on a sufficiently large window of the past will effectively
perform batch updates based on a set of trajectories sampled by different
``generators'', thereby performing meta-learning. 

\subsection{Capacity limitations}

In our analysis, we have assumed that the solution found is the minimizer
of the meta-learning objective. This is a very strong assumption. In
practice, the difficulty of actually finding a near-optimal solution
depends on many factors.

The first and most important factor is of course the model used for
regression. Properties such as e.g.~the inductive biases
of the model implementing the function class $\cal F$, 
the smoothness of the loss landscape, the optimizer,
and the memory capacity, play fundamental roles in any machine learning
method, and meta-learning is no exception. 

Another important factor is the task class, including both the space of hypotheses
and the loss function. Together they shape the complexity of the strategy
to regress via meta-learning. In particular, the invariances in the state machine (see Section~\ref{subsec:anatomy}) reduce the number of distinct mappings from past experiences $x_{<t}$ in $\cal{X}^\ast$ to instantaneous strategies $\pi_t$ in $\Pi$ the regressor has to learn. Conversely, in the worst case when there are no invariances, the number of distinct mappings grows exponentially in the maximum length~$T$ of the trajectories.

\subsection{Selected future challenges}

\paragraph{Task structure}
Meta-learning crucially relies on the skillful design 
of the class of tasks. Previous work has shown that
agents can meta-learn to perform a variety of task
if the task distributions have been designed accordingly,
such as learning to identify a best option \citep{denil2016learning,
wang2016learning} or learning to reason causally \citep{dasgupta2019causal}

In each case, the practitioner must ask the question: 
if the meta-learned strategy should have a
capability X, what property Y must the class of tasks posses? 
For instance, how should we design the generators so that we can 
generalize out-of-distribution and beyond the length of 
the trajectories sampled during meta-learning? 

Addressing questions like these entail further questions regarding the structure of tasks; but to date, we are not aware of an adequate language or formalism to
conceptualize this structure rigorously. In particular, we expect to 
gain: a better understanding of the dynamical structure of solutions;
predict the structure of the sufficient statistics that a class of
tasks gives rise to; and compare two tasks classes and determine if they 
are equivalent (or similar) in some sense.

\paragraph{Beyond expected losses} 
In the basic meta-learning scheme the strategies minimize the \emph{expected} loss. Minimizing an expectation disregards the higher-order moments of the loss distribution (e.g.\ variance), leading to risk-insensitive strategies that are brittle under model uncertainty. Going beyond expected losses means that we have to change our \emph{certainty-equivalent}---that is, the way we aggregate uncertain losses into a single value \citep{hansen2008robustness}.

Changing the certainty-equivalent changes the attitude towards risk. This has important applications ranging from safety \& robustness \citep{van2010risk, ortega2018modeling} to games \& multi-agent tasks \citep{mckelvey1995quantal, mckelvey1998quantal, littman1994markov}. For instance, if the agent is trained on an imperfect simulator of the real-world, we would want the agent to explore the world cautiously.

Risk-sensitive strategies can be meta-learned by tweaking 
the statistics of the generative process. For instance, by changing the prior over generators as a function of the strategy's performance, we can meta-learn strategies that are risk-sensitive. For games, it might be necessary to simultaneously meta-train multiple agents in order to find the equilibrium strategies.

\paragraph{Continual learning}

The continual learning problem asks an agent to learn 
sequentially, incorporating skills and knowledge in a 
non-disruptive way. There are strong links between 
the continual learning problem and the meta-learning 
problem. 

On one hand, the traditional meta-learning setting assumes a fixed 
distribution of tasks, which can be 
restrictive and unrealistic.  Allowing the distribution 
to change over time not only would be crucial from a 
practical perspective, but could also be used as a tool 
to refine the task distribution in order to induce 
the right properties in the learned solution. Continual 
or incremental variants of the meta-learning problem 
is however an under-explored topic (e.g. \citet{nagabandi2018deep} 
touches on this topic).

On the other hand, meta-learning can be seen as part 
of the solution for the continual learning problem. 
In principle continual learning is ill-defined. Remembering 
a potentially infinite set of skills is unfeasible 
within a finite model. While continual learning usually 
focuses on 0-shot transfer \citep{GoodfellowMDCB13,
ParisiKPKW18,KirkpatrickPRVDRKQTGDCDH16}, or 
how well the agent remembers a previously seen task 
without any adaptation, this might be the wrong measure.
A relaxation of this approach, e.g.\ explored by 
\citet{KaplanisSC18}, would be to measure how fast one 
recovers performance, 
which converts continual learning into a meta-learning 
problem. The compression of all seen tasks becomes 
the meta-learning algorithm that the agent needs to 
infer and that can be exploited to recover the solution 
of the task. However the tasks are not seen necessarily 
in an i.i.d.\ fashion. So while the mechanism outlined 
in this work could describe such a solution to a 
continual learning problem, it is unclear how to
approach the learning problem in practice.

\subsection{Conclusions}

Reinforcement learning algorithms based on probabilistic
models promise to address many of the shortcomings---in particular,
the sample-inefficiency---of model-free reinforcement learning 
approaches. However, the implementation of such systems is
very challenging and, more often than not, depends 
on domain expertise, i.e.\ human knowledge, hand-crafted in the 
form of probabilistic models that are either tractable but 
too simple to be useful, or outright intractable. Such
an approach does not scale.

Memory-based meta-learning offers a conceptually simple 
alternative for the construction of agents implicitly based 
on probabilistic models that leverages data and large-scale 
computation. In essence, meta-learning
transforms the hard problem of probabilistic
inference into a curve fitting problem. Here we have
provided three meta-learning templates: one for building predictors,
one for Thompson samplers, and one for Bayes-optimal agents 
respectively. 
In all of them, the key idea is to precondition a slow-learning
system by exposing it to a distribution over trajectories
drawn from a broad class of tasks, so that the meta-learned 
system ends up implementing a general and fast 
sequential strategy---that is, a strategy with the right 
inductive biases.

We have also shown why this approach works 
and how the meta-learned strategies are implemented. Basically,
the resulting strategies are near-optimal by construction 
because meta-learning directly trains on a Monte-Carlo 
approximation of the expected cost over possible hypotheses.
The sequential data drawn during this Monte-Carlo approximation 
is implicitly Bayes-filtered, and the Bayesian updates are 
amortized by the meta-learned strategy. Moreover, we have
shown that the adaptation strategy is implemented as a state
machine in the agent's memory dynamics, which is driven by
the data the agent experiences. A given memory state then
represents the agent's information state, and, more precisely, the
sufficient statistics for predicting its future interactions.
The structure of the transition graph encodes the symmetries
in the task distribution: paths that have the same initial
and final information state are indistinguishable, and thus 
equivalent for the purposes of modulating the future behavior 
of the agent.

Finally, we note that meta-learning also converts complex probabilistic inference problems into regression problems in \textit{one-shot} settings, when meta-learning is applied without memory (e.g. \citeauthor{orhan2017efficient} \citeyear{orhan2017efficient}). A key distinction of the sequential setting with memory is that meta-learning also produces an update rule. 

Our hope is that readers will find these insights useful,
and that they will use the ideas presented here as a conceptual 
starting point for the development of more advanced algorithms 
and theoretical investigations. There are many remaining 
challenges for the future, such as understanding task structure, 
dealing with out-of-distribution
generalization, and continual learning (to mention some). 
Some of these can be addressed through tweaking the basic 
meta-learning training process and through designing hypothesis 
classes with special properties.

More generally though, memory-based meta-learning illustrates
a more powerful claim: a slow learning system, given enough data 
and computation, can not only learn a model over its environment, 
\emph{but an entire reasoning procedure}. In a sense, this suggests
that rationality principles are not necessarily pre-existent, 
but rather emerge over time as a consequence of the situatedness 
of the system and its interaction with the environment. 

\subsection*{Acknowledgements}
We would like to thank Shakir Mohamed for his very thoughtful
feedback on the manuscript.

\subsection*{Changes}
\begin{description}
 \item[V1.0:] (08/05/2019) Original version.
 \item[V1.1:] (18/07/2019) Added a new reference; corrected error in 
 Figure~\ref{fig:minimal-state-machine}; added acknowledgements.
\end{description} 

\bibliographystyle{plainnat}
\bibliography{bibliography}

\end{document}